\documentclass{article}
\pdfoutput=1 

\usepackage{neurips_data_2022}

\usepackage[utf8]{inputenc} 
\usepackage[T1]{fontenc}    
\usepackage{hyperref}       
\usepackage{url}            
\usepackage{booktabs}       
\usepackage{amsfonts}       
\usepackage{nicefrac}       
\usepackage{microtype}      
\usepackage{xcolor}         
\usepackage{graphicx}
\usepackage{pgfplots}
\usepackage{color-edits}

\addauthor{MM}{purple} 

\title{Constructing the CORD-19 Vaccine Dataset}

\author{%
  Manisha Singh \\
  University of Washington \\
  \texttt{manishas@uw.edu}
  \And
  Divy Sharma \\
  University of Washington \\
  \texttt{divy@uw.edu} \\
  \AND
  Alonso Ma \\
  University of Washington \\
  \texttt{amatake@uw.edu} \\
  \And
  Bridget Tyree \\
  University of Washington \\
  \texttt{btyree@uw.edu} \\
  \And
  Margaret Mitchell \\
  University of Washington \\
  \texttt{margarmitchell@gmail.com} \\
}

\begin{document}
\nolinenumbers

\maketitle

\begin{abstract}
  We introduce  new dataset ‘CORD-19-Vaccination’\footnote{Our dataset is available at \url{https://github.com/manisha-Singh-UW/CORD-19-Vaccination}} to cater to scientists specifically looking into COVID-19 vaccine-related research. This dataset is extracted from CORD-19 dataset \citep{wang-etal-2020-cord} and augmented with new columns for language detail, author demography, keywords, and topic per paper. Facebook’s fastText model is used to identify languages \citep{https://doi.org/10.48550/arxiv.1612.03651}. To establish author demography (author affiliation, lab/institution location, and lab/institution country columns) we processed the JSON file for each paper and then further enhanced using Google's search API to determine country values. ‘Yake’ was used to extract keywords from the title, abstract, and body of each paper and the LDA (Latent Dirichlet Allocation) algorithm was used to add topic information \citep{CAMPOS2020257, 10.1007/978-3-319-76941-7_63, 10.1007/978-3-319-76941-7_80}. To evaluate the dataset, we demonstrate a question-answering task like the one used in the CORD-19 Kaggle challenge \citep{cord19kaggle}. For further evaluation, sequential sentence classification was performed on each paper’s abstract using the model from \citet{https://doi.org/10.48550/arxiv.1612.05251}. We partially hand-annotated the training dataset and used a pre-trained BERT-PubMed layer. ‘CORD-19-Vaccination’ contains 30k research papers and can be immensely valuable for NLP research such as text mining, information extraction, and question answering, specific to the domain of COVID-19 vaccine research. 
\end{abstract}

\section{Introduction}
\label{intro}
A report released in early 2021 declared, “World to spend \$157 billion on COVID-19 vaccines through 2025” \citep{157billion}. Despite this, there are no datasets that are specific to COVID-19 vaccine research. The COVID-19 Open Research Dataset (CORD-19) \citep{wang-etal-2020-cord} is a corpus of academic papers on coronavirus research. However, the metadata file for the CORD-19 dataset (release version 109) consists of over one million journals, resulting in big data issues and information overload. The overall goal is to create a dataset that was based out of CORD-19 but only includes the papers that are relevant to vaccine research. In this work, we introduce a dataset curated from the CORD-19 dataset and tailored to aid research on COVID-19 vaccines.

Our approach utilizes a pipeline of \textit{information extraction}, \textit{data augmentation}, and \textit{task implementation}:

\textbf{Extraction phase: } In this phase, we created a SQLite data pipeline to manage the large volume of the CORD-19 dataset. The language of each paper’s abstract was determined using Facebook’s fastText library \citep{https://doi.org/10.48550/arxiv.1612.03651}. Subsequently, using SQLite query we created a subset of the CORD-19 dataset, taking only those papers where the starting ‘publish time’ was ‘2020’ and either the ‘Abstract’ or ‘Title’ contained the word ‘vaccine’ or ‘vaccination’ in all the languages present in CORD-19. 

\textbf{Data augmentation phase: } In this phase we added new columns to the dataset. The language ID determined from the previous phase was retained. Data on author affiliation was collected from the ‘json parse’ files of the research papers web search of each research paper. Keywords were added using ‘Yake’ \citep{CAMPOS2020257}. Finally, we implemented ‘Topic modeling’ where we classified the dataset into topics based on the ‘Abstract’ using the LDA model \citep{https://doi.org/10.48550/arxiv.1612.05251}.  

\textbf{Task implementation phase: } We implemented ‘Question and Answering’ and ‘Sequence sentence classification’ task using the CORD-19-Vaccination dataset.  

\begin{figure}[htp]
    \centering
    \includegraphics[width=\textwidth]{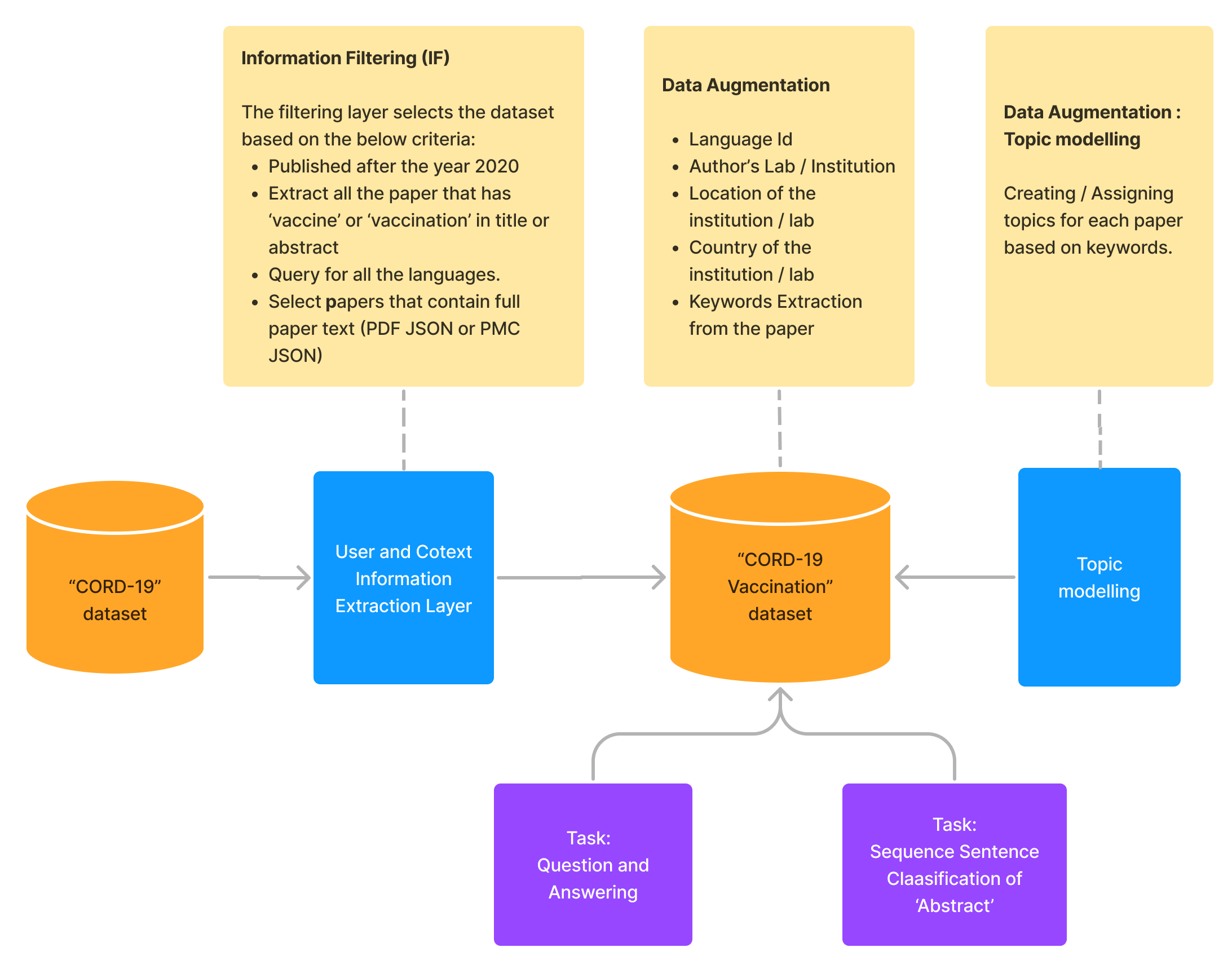}
    \caption{CORD-19-Vaccination dataset creation - overview}
    \label{fig:cord-19-vaccination-creation}
\end{figure}

The implementation of each of these steps is detailed in the sections below. Figure \ref{fig:cord-19-vaccination-creation} shows a visual overview of the creation of the dataset.

In what follows, we motivate and describe each phase, highlighting discoveries at each step. We then demonstrate the utility of the dataset on the tasks of Question and Answering and Sequential Sentence Classification.

\section{Extraction phase: user and context information extraction from CORD-19}
\label{extraction}
\begin{figure}[htp]
    \centering
    \includegraphics[width=12cm]{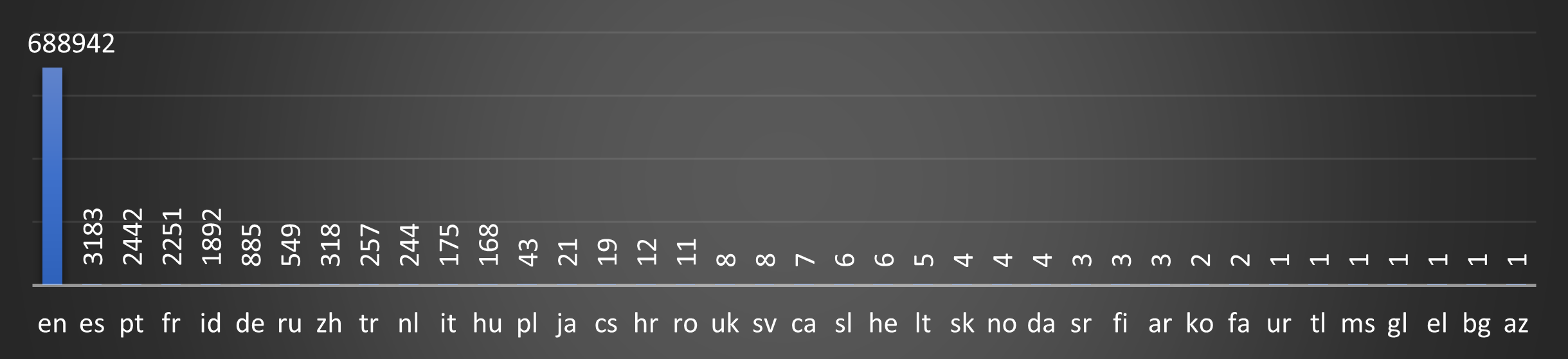}
    \caption{Language distribution in CORD-19}
    \label{fig:lang_id_cord_19}
\end{figure}

The CORD-19-Vaccination dataset is extracted from the CORD-19 dataset based on the following filtering criteria:  
\paragraph{Publish time:}
The CORD-19  metadata.csv has a column \texttt{publish\_time}. The extraction filter extracts all data where \texttt{publish\_time} is greater than or equal to '2020'. 
\paragraph{Pattern search:}
'vaccine'/'vaccination' in either \texttt{title} or \texttt{abstract}: CORD-19 metadata.csv does not indicate the language of a paper. Figure \ref{fig:lang_id_cord_19} shows the language distribution of CORD-19 using the fastText model. CORD-19-Vaccination dataset is extracted from the CORD-19 dataset. In order to extract papers with the word 'vaccine/vaccination' in every language a query for information extraction was customized to search the pattern of 'vaccine/vaccination' in every language. This query was applied on columns \texttt{title} and \texttt{abstract}. The language ID abbreviations are taken from \citet{istdlanguagestudy}. 

\begin{figure}[htp]
    \centering
    \includegraphics[width=6cm]{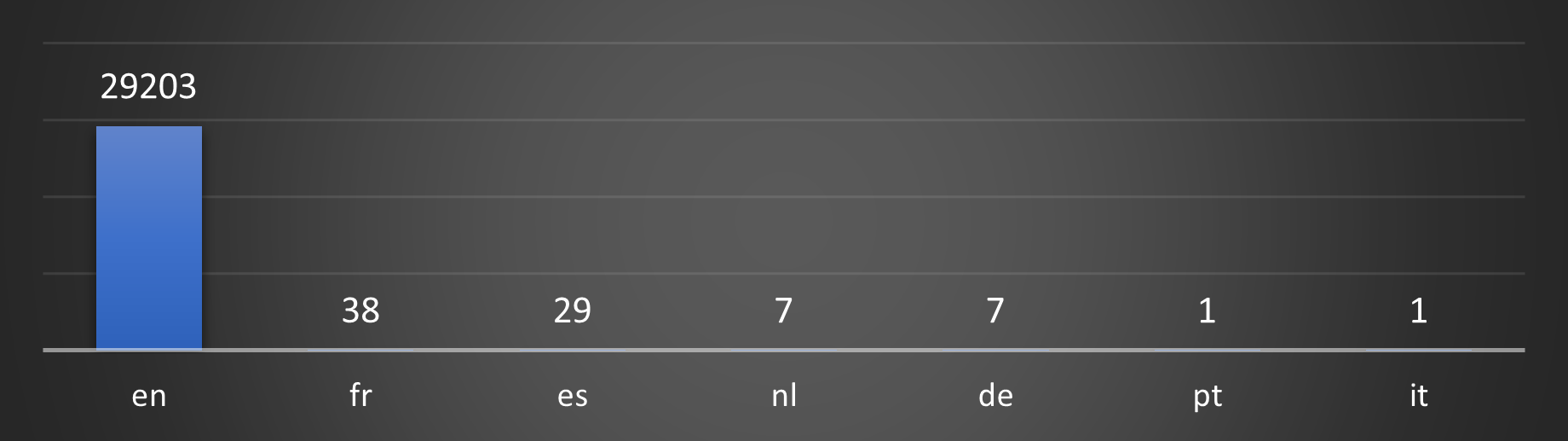}
    \caption{Language distribution in CORD-19-Vaccination}
    \label{fig:lang_id_cord_19_vacc}
\end{figure}

\paragraph{Pdf\_json\_files / pmc\_json\_files:}
CORD-19 metdata.csv has the columns \texttt{pdf\_json\_files} and \texttt{pmc\_json\_files}. These columns give the path of the json files in the CORD-19 dataset. All papers selected had the \texttt{pdf\_json\_file} or \texttt{pmc\_json\_files}  present.

\paragraph{Abstract is not null:}
All papers selected had the \texttt{abstract} column present. Our exploratory data analysis revealed that almost all standard published papers must follow a particular template where the abstract must be present. The papers which did not have an abstract were mostly articles that were not published papers. This improved the quality of our dataset as it only includes research papers.  

Since several of the ‘CORD-19-Vaccination’ columns are based on CORD-19, the preprocessing for CORD-19 is detailed in \citet{wang-etal-2020-cord}. 

\section{Data augmentation phase: language detail, author demography, keyword and topic}
\label{augmentation}

Table \ref{table:augmentedfields} includes a list of fields added to the CORD-19-Vaccination dataset. The code to generate these augmented fields is available in the GitHub repo\footnote{The code for the data augmentation is available at \url{https://github.com/manisha-Singh-UW/CORD-19-Vaccination}}.
\begin{table}[htp]
\caption{CORD-19-Vaccination augmented fields}
\label{table:augmentedfields}
\centering
\begin{tabular}
{p{.25\textwidth}p{.25\textwidth}p{.2\textwidth}p{.15\textwidth}}
\hline
\textbf{Language Detail}     &  \textbf{Author Demography} & \textbf{Keywords} & \textbf{Topic}  \\\hline
\texttt{lang\_id},  \texttt{lang\_id\_confidence}, \texttt{lang\_id\_predictions}    & \texttt{aff\_lab\_inst}, \texttt{aff\_location}, \texttt{aff\_country} & \texttt{keywords}  & \texttt{topic}, \texttt{topic\_index}, \texttt{topic\_prob}  \\ \hline
\end{tabular}
\end{table}

\subsection{Language id}
Language ID is included in the dataset in order to support text demography. The input to the fastText model is the text of the `abstract' from each paper and the output are the three fields as shown in  Table \ref{table:langidsample}. 
\begin{table}[htp]
\caption{Language id - sample data}
\label{table:langidsample}
\centering
\begin{tabular}{lll}
\hline
\textbf{lang\_id} & \textbf{lang\_id\_confidence} & \textbf{lang\_id\_predictions}           \\ \hline
en       & 0.9167               & en=0.9167, id=0.0055, fr=0.0043 \\ \hline
\end{tabular}
\end{table}
The fastText model predicts the language as 'English' with a confidence level of '0.9167'. However, the fastText model also gives a small confidence level to 'Indonesian' at '0.0055' and 'French' at '0.0043'. This is likely due to the medical domain including many loan words. In the example in Table \ref{table:langidsample}, the confidence level of ‘English’ compared to the other languages is much higher, so Language ID field is set to ‘English’. The graph in Figure \ref{fig:lang_id_cord_19_vacc} shows the language distribution of the CORD-19-Vaccination dataset. 

\subsection{Author's demography (lab/institution location and country)}
The CORD-19 dataset contains authors and a journal name for each paper.  However, in order to get more details regarding the authors' demography, we augmented the data with authors' `lab/institution affiliation',  `lab/institution location' and `lab/institution country'. Details on the authors' demography can be used to construct a collaboration network to illustrate collaborations or coauthor-ship relations among institutions as in the article \citet{demystifying}. 

\begin{table}[htp]
\caption{Author detail - sample data }
\label{authordetail}
\centering
\begin{tabular}{lll}
\hline
\textbf{aff\_lab\_inst}                                                                       & \textbf{aff\_location}                                                                              & \textbf{aff\_country} \\ \hline
\begin{tabular}[c]{@{}l@{}}University of Maryland \\ School of Medicine\end{tabular} & \begin{tabular}[c]{@{}l@{}}postCode=21201; region=MD; \\ settlement=Baltimore\end{tabular} & USA          \\ \hline
\end{tabular}
\end{table}

The author's 'lab/institution affiliation',  'lab/institution location'  and 'lab/institution country' was not mentioned in the CORD-19 metadata file. However, in order to do descriptive analysis such as number of the papers contributed by each institution, geographical distribution of institutions and collaboration among institutions from different countries/regions we need the institution details related to each author of the paper.  

 Additionally, as the country of affiliation metadata was only available for approximately 63\% of the JSON files, further data augmentation was carried out to extract the country of the first author via web scraping. For this process, titles of papers with missing country data were searched through Python’s Google search API and the HTML source code of the webpage corresponding to the first query result was parsed using Selenium and Beautiful Soup. Scraped titles from the search query and their linked countries of affiliation were stored and subsequently validated by comparing similarity between the original CORD-19 paper title and the scraped title. Entries with a similarity below 0.4 (calculated using the Sequence Matcher module from Python’s diff lib library) were excluded. 

\paragraph{Author demographic:}
Speaker/Author demographic was mainly assessed via examination of the distribution of first author’s countries of affiliation. We initially extracted the country data from the full text JSON files, achieving coverage of 63\% over the total of papers. Through web scraping, we identified the country of affiliation for an additional group of papers, increasing coverage to 93\%.

\begin{figure}[htp]
    \centering
    \includegraphics[width=10cm]{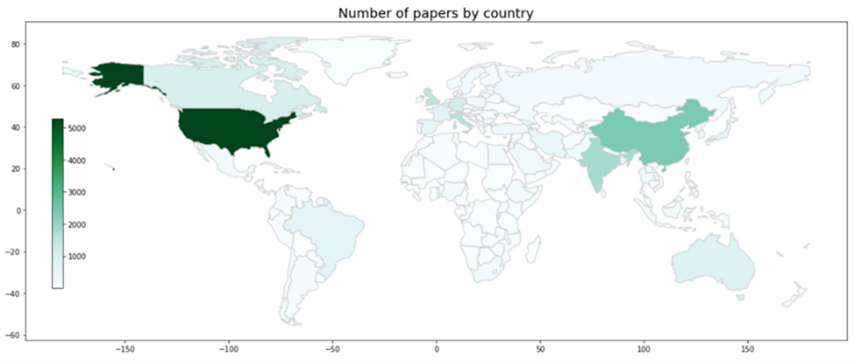}
    \caption{Number of papers by first author country of affiliation}
    \label{fig:country_affiliation}
\end{figure}

50\% of the total papers were concentrated over 7 countries: United States of America, China, India, Italy, United Kingdom, Germany, and Canada, with the USA representing 20\% of the dataset. A complete map depicting the distribution of number of papers by country of affiliation of the first author can be observed in Figure \ref{fig:country_affiliation}. Most notably, apart from the concentration of research in the previously mentioned countries, a stark lack of representation from the Global South (with the exception of Brazil) was also evident.  

The authors' detail is present in the associated JSON file of each paper, from which institution of affiliation, location and country were extracted. The input is the JSON file and output are the columns corresponding to each author and paper id.

\subsection{Keywords from 'abstract', 'title' and 'text body'}
Top 20 keywords have been extracted using the Yake library. Keywords from every paper can be used further in topic modeling and for keyword search, which gives an idea about the main content of the paper. 

Yake was used because it uses an unsupervised approach, which is corpus-independent, and domain and language independent. Yake follows an unsupervised approach that builds upon features extracted from the text, making it applicable to documents written in different languages without the need for domain-specific knowledge.  

The input to the Yake object is the text string, generated from the 'title', 'abstract' and 'body text' of the paper. The output is the list of the keywords. One can customize the number of the top key words and n-grams. For our current implementation we have chosen the top-20 keywords and n-grams size as '3'. The rest of the parameters for Yake are default values. A sample result of the keyword extraction is shown in Table \ref{table:keywords}. 

\begin{table}[htp]
\centering
\caption{Sample data: keywords of a paper}
\label{table:keywords}
\begin{tabular}{l}
\hline
\multicolumn{1}{c}{\textbf{Keywords}}                          \\ \hline
\begin{tabular}[c]{@{}l@{}}DNA vaccine; archaeosome; DNA; recombinant DNA vaccine; \\ pDNA - surface localized archaeosome ; archaeosome vaccines group; \\ cells; DNA vaccine candidate; localized archaeosome; \\ vaccine; archaeosome vaccines; groups; \\ plasmid DNA; gene DNA vaccine; PBS control groups; \\ recombinant gene; pDNA-encapsulated archaeosomes; gene; mice; control groups\end{tabular} \\ \hline
\end{tabular}
\end{table}

\subsection{Topic modeling}
We further augmented the dataset by implementing a topic modelling algorithm (Latent Dirichlet Allocation). Generation of topic labels has a twofold intention: first, given the variety of possible themes within the papers (even when filtered to only include vaccine-related documents), it provides a comprehensive overview of recurrent subjects and allows for easy inspection of the distribution amongst them. Second, it facilitates quick sub-setting of the data to allow potential users to fit more scoped tasks.

Training of the LDA was performed over the complete set of paper abstracts, which were pre-processed into lower case; and had stop words, punctuations, and small words (e.g. character length below 3) removed. We tested a range of “number of topics (n)” parameters (from n=5 to n=14) and evaluated each model via the Coherence score described by \citet{10.1145/2684822.2685324}, ultimately selecting n=5 as the final parameter due to its higher score and parsimony.

To assign a label to each trained topic, we selected the top 20 words per topic and cross-referenced this list against the paper titles of the documents with the highest probability of pertaining to a particular theme. Additional evaluation was performed visually to assess possible topic overlaps by applying dimensionality reduction (through t-SNE) over the topic distribution vectors of each document and plotting them, colored by label. The resulting topic labels and their distribution over the dataset are shown in Table \ref{table:topicmodel}. 

\begin{table}[htp]
\centering
\caption{Topic distribution in CORD-19-Vaccination dataset }
\label{table:topicmodel}
\begin{tabular}{ll}
\hline
\textbf{Topic}                                          & \textbf{\% of dataset} \\ \hline
T1: Vaccine development                                 & 20\%                   \\ 
T2: Vaccination side-effects                            & 14\%                   \\ 
T3: Vaccination efficacy                                & 16\%                   \\ 
T4: Methodologies for COVID studies (e.g. simulations)  & 25\%                   \\ 
T5: Vaccine uptake (by factors of age, sex, race, etc.) & 25\%                   \\ \hline
\end{tabular}
\end{table}

\section{Task implementation phase}
CORD-19-Vaccination dataset contains the metadata of approximately 30k research papers. As the next step, we evaluated the dataset by performing ‘Question and Answering’ and ‘Sequential Sentence Classification’ tasks. 
\subsection{Question and answering task}
We designed a task similar to the \citet{cord19kaggle}'s Kaggle competition challenge on CORD-19. 'Covid-19 vaccine' Question and Answering system is a domain specific task. In an ideal situation we need a medical expert to design the questions and evaluate the answers. However, in absence of a medical expert we designed a simple vaccine specific question. We tried to follow the "user-based approach" as per \citet{diekema} to evaluate the answer.

The Question and Answering task consists of three parts: 'question', 'context', and 'answer'. The input to the model is a covid-19 vaccine specific question and the context. In this implementation we are assuming that the question is contained in the context. We needed to keep the context small to implement this model on 30k papers. This is done by selecting the papers similar to the answers, using 'Okapi BM25' \citep{okapiwiki}. Okapi BM25 is a ranking function used by search engines to estimate the relevance of the document for a given search query. For each question and context, we are using “Huggingface transformer library” to predict the answer \citep{wolf-etal-2020-transformers}. We have used the pretrained QA model 'bert-large-uncased-whole-word-masking finetuned-squad'. The solution for this task was customized for ‘CORD-19-vaccination’ dataset which is inspired by \citet{qamodel}'s Kaggle notebook.  

\begin{figure}[htp]
    \centering
    \includegraphics[width=12cm]{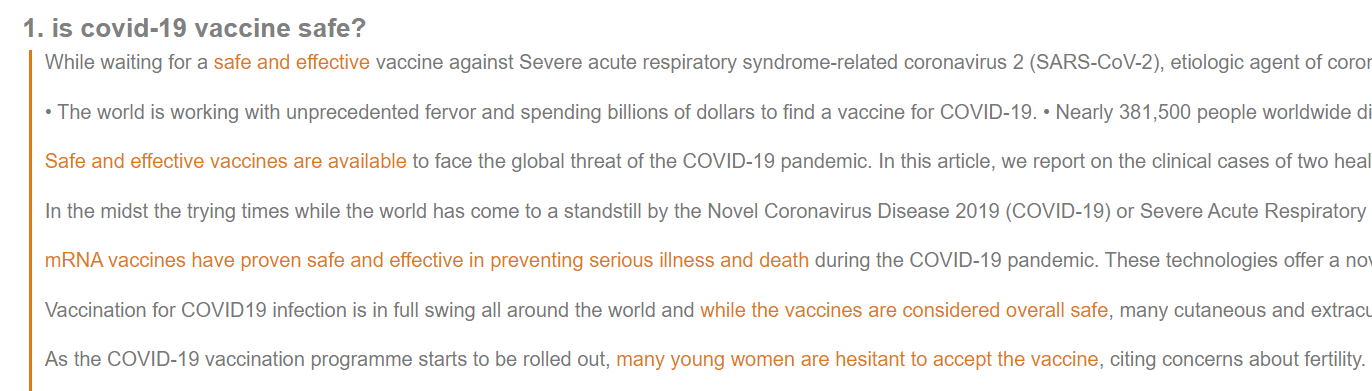}
    \caption{Question and answering output}
    \label{fig:qanda}
\end{figure}

Figure \ref{fig:qanda} is output of the ‘Question and Answering’ task for the question ‘is covid-19 vaccine safe?’. 

\begin{table}[htp]
\caption{Answers evaluation: 'user-based' analysis }
\label{table:q&atable}
\begin{tabular}{lllll}
\hline
\textbf{Question}                  & \textbf{Papers}                     & \textbf{Citation} & \textbf{Viewed} & \textbf{Downloads} \\ \hline
Is Covid-19 vaccine safe? & 10.1038/s41577-021-00525-y & 111      &        &           \\ 
                          & 10.1016/j.puhe.2020.05.007 & 9        &        &           \\ 
                          & 10.1093/jlb/lsaa024        & 3        & 1146   & 435       \\ 
                          & 10.3390/vaccines10020298   &          & 627    &           \\ 
                          & 10.1111/jdv.17499          & 3        &        &           \\ 
                          & 10.1111/dth.15146          & 6        &        &           \\ \hline
\end{tabular}
\end{table}
Table \ref{table:q&atable} gives the list of the papers as answers for the question 'Is Covid-19 vaccine safe? '. According to the "user-based approach" of evaluation we can say that the papers in the result seem relevant, as most of the papers were recently published.

'CORD-19-Vaccination' dataset is better than 'CORD-19' for vaccination related 'Question and Answering' task due to following reasons: 'CORD-19-Vaccination' was augmented with fields in Table \ref{table:topicmodel}. These fields are not present in 'CORD-19'. Researchers using  'CORD-19-Vaccination' can make use of these augmented fields for better answers. 

The column \texttt{keyword} in 'CORD-19-Vaccination' dataset is the list of keywords extracted from the body of the text papers using 'Yake', so if we extract the context search using 'Title'/'Abstract'/'Keyword', the answer should be more accurate. 

\subsection{Sequential sentence classification task}
Text classification is a very important task in Natural Language Processing (NLP) where a label or class is assigned to a text. In the current task, the focus is on the classification of sentences in medical abstracts. The sentences in the abstracts appear in a sequence therefore this task is called "Sequential Sentence Classification Task". This task converts unstructured block-of-text abstracts into structured abstracts (text organized into semantic headings such as Background, Methods, Results, and Conclusions), making it easy to quickly locate relevant information. This task is based on the paper \citet{https://doi.org/10.48550/arxiv.1612.05251}. The output of this task has been provided as a new column named \texttt{labeled\_abstract} in the dataset.

The data for training the model for this task is obtained from the PubMed 200k RCT dataset \citet{DBLP:journals/corr/abs-1710-06071} and the CORD-19 dataset itself. 11.58\% of the abstracts from the CORD-19 dataset (approximately 117k samples) were found to have abstracts structured with semantic headings. Similarly, 14.66\% of the records in the CORD-19-Vaccine dataset (4294 samples) were found to have abstracts structured with semantic headings. These records were split into test and validation datasets for model training. A single data sample contains information on target labels, sentence from abstract, and order of sentences, compatible with \citet{https://doi.org/10.48550/arxiv.1612.05251}. The \texttt{pubmed\_id} and \texttt{cord\_uid} fields are available as comments and are not inputs to the training model. As per the guidance of the PubMed 200k RCT paper, numbers from the dataset have been replaced with the @ sign.

\begin{figure}[htp]
    \centering
    \includegraphics[width=7cm]{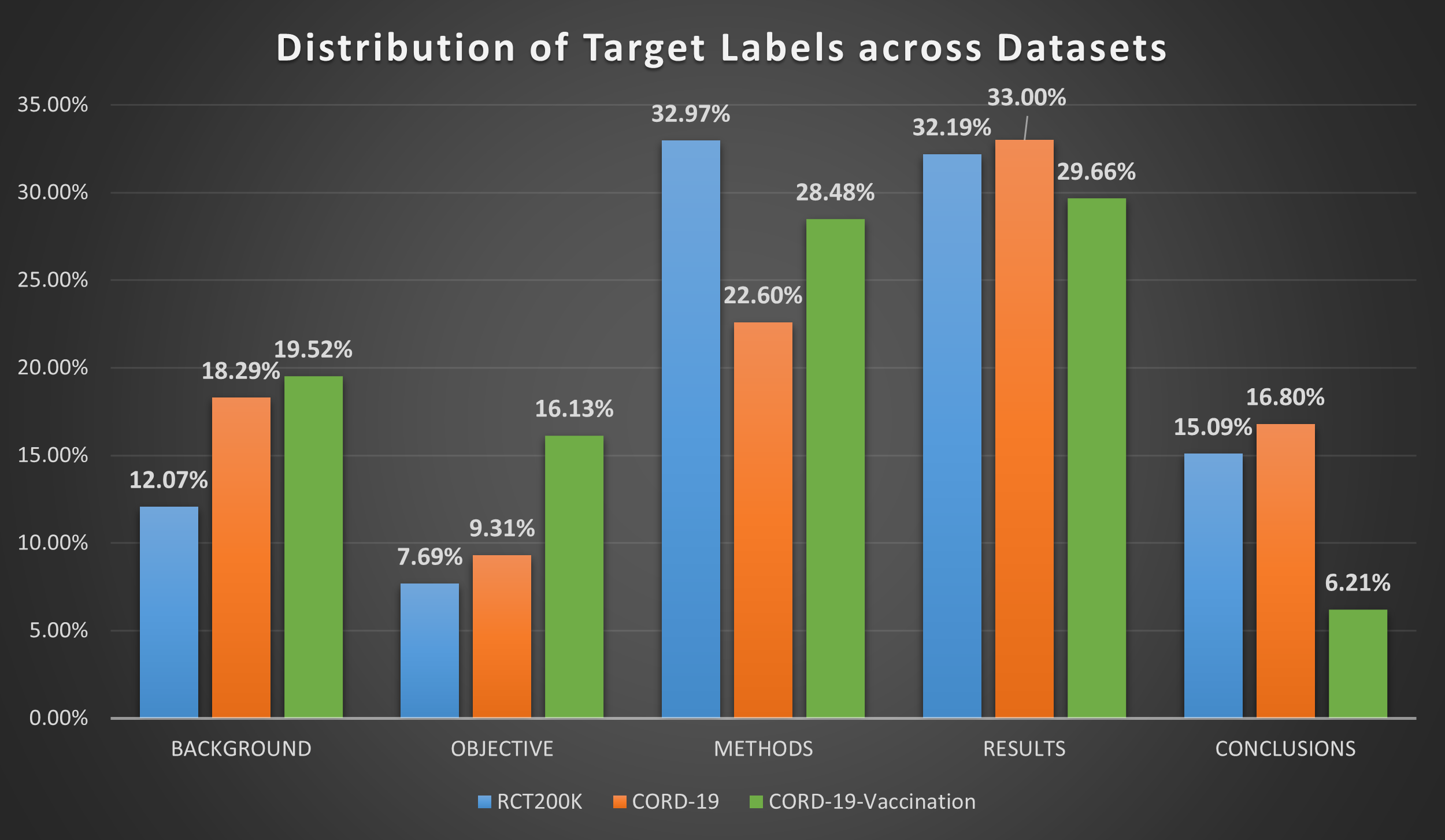}
    \caption{Distribution of target labels across datasets}
    \label{fig:targetlabels}
\end{figure}

The distribution of various target labels is shown across datasets in Figure \ref{fig:targetlabels}. It is important to note that the percentage of OBJECTIVE labels is quite high (at 16.13\%) in the CORD-19-Vaccination dataset, while the percentage of CONCLUSION labels is quite low (at 6.21\%) compared to PubMed RCT200k and CORD-19 datasets. 

\begin{figure}[htp]
    \centering
    \includegraphics[width=\textwidth]{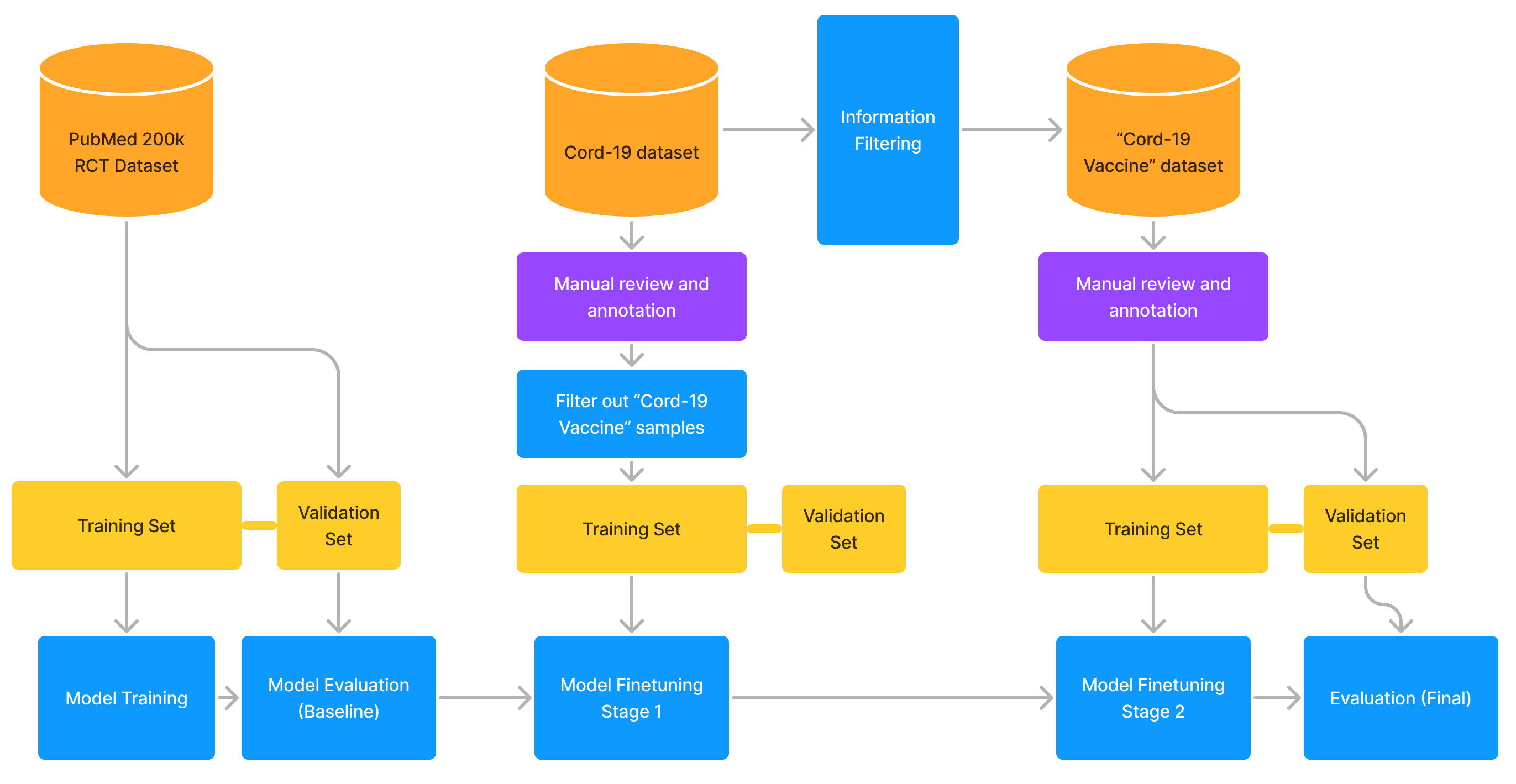}
    \caption{Task workflow}
    \label{fig:taskworflow}
\end{figure}

\paragraph{Task Workflow:} 
The workflow in figure \ref{fig:taskworflow} shows the sequence of tasks performed during the training and subsequent fine-tuning of the model. This particular workflow was chosen to allow coarse-grained to fine-grained model training. 

The model architecture used for training is based on the \citet{https://doi.org/10.48550/arxiv.1612.05251} paper. A pre-trained and frozen BERT-PubMed layer has been used to improve performance. The original training/validation data split of PubMed 200k RCT dataset was used fir the initial round of training. For fine tuning, a random split of 70-30 was used for stage 1 and split of 50-50 was used for stage 2. The model training was performed initially with a learning rate of 1e-4, which was reduced to 1e-4 for fine tuning. A system based on Nvidia Tesla P100 GPU was used for training.

\paragraph{Output:} 
Table \ref{table:performancemetric} shows performance metrics of this model on the CORD-19-Vaccine dataset.

\begin{table}[htp]
\centering
\caption{Performance metric on CORD-19-Vaccine}
\label{table:performancemetric}
\begin{tabular}{llll}
\hline
\textbf{Accuracy}         & \textbf{F1-score}           & \textbf{Precision}         & \textbf{Recall}            \\ \hline
0.7618 & 0.7569 & 0.7569 & 0.7618 \\ \hline
\end{tabular}
\end{table}

\begin{figure}[htp]
    \centering
    \begin{minipage}{0.45\textwidth}
    \centering
    \includegraphics[width=7cm]{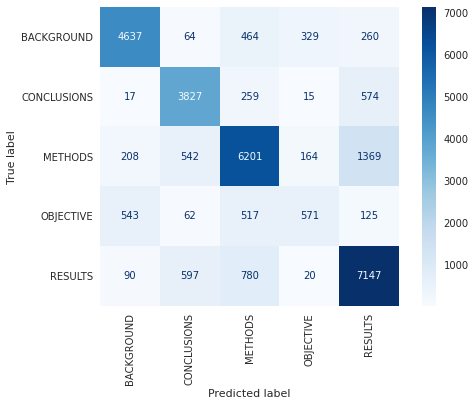}
    \caption{Confusion matrix on raw numbers}
    \label{fig:conmatmarg}
    \end{minipage}\hfill
    \begin{minipage}{0.45\textwidth}
    \centering
    \includegraphics[width=7cm]{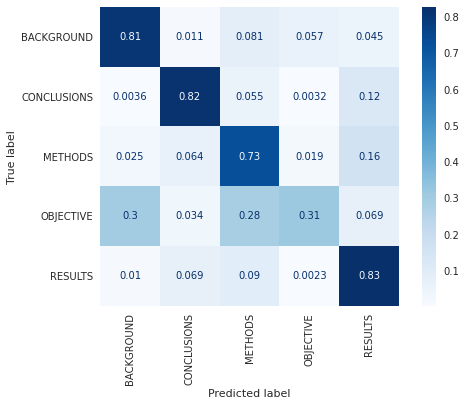}
    \caption{Confusion matrix on normalized true labels}
    \label{fig:conmatnorm}
    \end{minipage}
\end{figure}

\paragraph{Evaluation:} 
Figures \ref{fig:conmatmarg} and \ref{fig:conmatnorm} are the Confusion Matrix plotted using scikit learn. The matrix in Figure \ref{fig:conmatmarg} shows the raw numbers of label distribution, while the matrix in Figure \ref{fig:conmatnorm} is normalized on the “true” labels.

We can observe from the confusion matrix that the OBJECTIVE label was often confused with BACKGROUND and METHODS. Similarly, METHODS label was often confused with RESULTS. 

Additional evaluation was performed by manually reviewing the most wrong predictions. Some patterns that we found in these predictions are short sentences consisting of just a few words were incorrectly predicted, and ungrammatical or ambiguous sentences were misclassified. 

\section{License}
‘CORD-19-Vaccination’ dataset is extracted from ‘CORD-19’ dataset, so ‘CORD-19-Vaccination’ dataset also follows all the licenses\footnote{CORD-19 Dataset License:  \url{https://ai2-semanticscholar-cord-19.s3-us-west-2.amazonaws.com/2020-03-13/COVID.DATA.LIC.AGMT.pdf}} that are followed by ‘CORD-19’.

\section{Conclusion}
In this paper, we are introducing our new dataset ‘CORD-19-Vaccination’. This dataset consists of approximately 30k rows of metadata of scientific research papers, specific to the domain of COVID-19 vaccine research, making it the largest known curated resource in this domain. This dataset has been augmented with valuable details that extends the information present in the CORD-19 dataset. The ‘Question and Answering’ and ‘Sequential Sentence Classification’ evaluation results further highlights the value of this dataset for various NLP tasks. We hope that the release of this dataset can be immensely valuable to the COVID-19 vaccine-research community and used for NLP research such as text mining, information extraction, and question answering, specific to the domain of COVID-19 vaccine research. 
\bibliographystyle{plainnat}
\bibliography{default}


\end{document}